\documentclass[10pt, a4paper]{article}

\usepackage{amsmath}
\usepackage{listings}
\usepackage{subcaption}
\usepackage{placeins}
\usepackage{hyperref}
\usepackage[final]{lrec2026} 

\title{Fully Automated Identification of Lexical Alignment and Preference-Stage Shifts in Large Language Models}

\name{Thomas Stephan Juzek\thanks{\hspace{-0.65cm}
Code: \href{https://github.com/fsu-nlp/lexical-alignment-shifts}{github.com/fsu-nlp/lexical-alignment-shifts}. 
Correspondence: TSJ. 
Contributions: TSJ led conceptualisation/methodology, implementation, validation/analysis, and writing, assisted by XM and JH.}, 
Xiaoyang Ming, Jose A. Hernandez}
\address{Florida State University \\
         \{tjuzek, xm24b, jah22q\}@fsu.edu\\}

\abstract{
The language used by digital chat assistants such as ChatGPT can diverge from human expectations (misalignment). Research, mostly on Scientific English, has described both \textit{what} divergences occur and, to some extent, \textit{why}, linking them to the training stage of human preference learning. Yet, existing approaches rely on manual curation. This paper introduces two curation-free, assumption-light evaluation metrics:\ the Lexical Alignment Score, which identifies lexical overuse, and the Triangulated Preference Shift, which quantifies how much of such shifts can be attributed to human preference learning. Using PubMed abstracts, continuations were generated and measured using windowed document prevalence across six model families (Falcon, Gemma, Llama, Mistral, OLMo, Yi). The procedure identifies, without manual intervention, overused items such as \emph{suggest}, \emph{additionally}, and \emph{strategy}, and estimates their link to preference learning. Our findings replicate prior work and remain stable across parameter settings, random seeds, and evaluation on further data. The approach scales readily and enables systematic study of lexical (mis)alignment beyond Scientific English and across languages, and as such, the metrics have the potential to contribute to improved alignment for future models and understanding of its origins. 
\\ \newline \Keywords{lexical alignment, evaluation, Large Language Models, preference learning, Scientific English} }

\begin{document}

\maketitleabstract

\section{Introduction}
\label{sec:intro}

Usage of digital assistants based on Large Language Models (LLMs), such as ChatGPT, is increasing fast, and these artificial intelligence (AI) tools are now widely used for programming, language editing, and information finding \cite{stackoverflow2024-ai,coffey2024-ihe-oup,apnews2025-apnorc-usage,pew2025-chatgpt-usage}. They perform strongly on standard benchmarks (e.g., legal, mathematical, and language tasks; \citealp{achiam2023gpt,hendrycks2020measuring,cobbe2021training}). Yet they can deviate from human usage in systematic ways. A notable instance of such \emph{misalignment} concerns lexical behaviour:\ in Scientific English, assistants disproportionately favour items such as ``delve'', ``intricate'', and ``furthermore'' \cite{matsui2024delving,kobak2024delving,liang2024mapping,liu2024towards,gray2024chatgpt,geng2024chatgpt}. Current evidence implicates the preference-learning stage -- e.g., reinforcement learning from human feedback (RLHF; \citealp{ouyang2022training}) or direct preference optmization (DPO; \citealp{rafailov2024direct}) -- as a notable driver of these shifts \cite{bharadwaj2025flattery,juzek2025word}.

Training of LLM-based chat assistants typically follows four stages:\ pre-training (giving base models), instruction tuning (to make models more helpful assistants), preference learning (aligning with human judgements), and task-specific fine-tuning \cite{christiano2017drlhf,wei2021finetuned,touvron2023llama2}. For models after instruction and preference tuning, we use \textit{instruct} models throughout the paper; they are sometimes referred to as \textit{chat} models. Importantly, preference learning has delivered substantial gains in assistant behaviour \cite{ouyang2022training,rafailov2024direct}.

However, evaluations of AI-associated lexical items, to both the \textit{what} (model-human misalignment; e.g.\ \citealp{matsui2024delving,kobak2024delving,liang2024mapping,liu2024towards,geng2024chatgpt}) and the \textit{why} (stage-specific shifts; e.g.\ \citealp{juzek2025word}), face a major gap:\ dominant approaches rely on manual curation and filtering (mainly ad hoc heuristics), which limits reproducibility and scalability across domains and languages. 

We introduce a curation-free evaluation pipeline over model generations. We treat lexical overuse as a tractable behavioural probe, and the main contribution is a diagnostic within the alignment pipeline, with pointers to attribution (causes for the observed model behaviour). For this, we evaluate six model families, with both base and instruct variants, using 42{,}000 PubMed abstracts as input. All generations use deterministic decoding, and symmetric cleaning is applied to human and model texts alike (Section~\ref{sec:task-and-data}). We then introduce two metrics:\ the \emph{Lexical Alignment Score} (LAS), which quantifies \emph{what} is overused relative to human continuations (Section~\ref{sec:las}); and the \emph{Triangulated Preference Shift} (TPS), which isolates the \emph{why}, that is, uplifts attributable to the preference-learning stage (Section~\ref{sec:tps}). The metrics give promising results, both for individual lexical items and for estimating how much of the observed shifts stem from preference learning, as well as macro-level model trends (Section~\ref{sec:results}). We validate the metrics by analysing additional data, varying parameters, and comparing results against prior literature (Section~\ref{sec:validation}). Our work, by providing more effective diagnostics, lays the foundations for more effective mitigation, which remains to be developed further. This situates our study alongside recent work suggesting that post-training can improve assistant behaviour while also narrowing output diversity, collapsing preference variation, or amplifying dominant response patterns \citep{kirk2023understanding, xiao2024algorithmic, zhang2025verbalized, murthy2025one}. Broader implications are discussed in Section~\ref{sec:discussion}.

\section{Related Work}
\label{sec:related}

\subsection{Lexical Overuse in LLMs} 

After the release of ChatGPT, a sudden spike in the usage of a small set of words (e.g., ``delve'', ``furthermore'', ``intricate'') in academic writing was noted \cite{matsui2024delving,kobak2024delving,liang2024mapping,liu2024towards}. As many of these words are also overused in AI-generated texts \cite{matsui2024delving}, the most plausible explanation for the spikes is that AI-assisted writing has contributed to these spikes \cite{kobak2024delving}. Whilst most work concentrates on Scientific English, there are parallel findings in journalism/newsroom contexts \cite{fitterer2025testing}. These analyses are largely based on written corpora; some work notes the rise of AI-associated language in unscripted speech \cite{yakura2024empirical,anderson2025model}, but causal attribution remains incomplete.

\subsection{Mechanisms Underlying Lexical Overuse}
Evidence implicates RLHF/DPO as \textit{a} driver of lexical overuse in Large Language Models. Triplet comparisons (Human vs Base vs Instruct) show patterns consistent with the hypothesis that preference learning plays a role \cite{bharadwaj2025flattery,juzek2025word}. For stylistic and formatting choices, small biases in preference data can produce disproportionately large behavioural changes \cite{zhang2024lists}. Work on self-training and synthetic instruction-tuning data highlights risks of feedback loops and style drift, which include possible output degradation \cite{alemohammad2023-mad,briesch2023-selfconsuming,shumailov2023-curse}. These findings further motivate automated diagnostics that can separate general misalignment from preference-stage effects.

\subsection{Limitations of Prior Identification Procedures}

Most procedures proposed in the literature to identify AI-overused words have two shortcomings (studies commonly show one or both; representative examples include \citealp{kobak2024delving,gray2024chatgpt,matsui2024delving,juzek2025word}):\ First, they rely on manual curation, particularly hand-tuned heuristics and post hoc filtering, which limits scalability, reproducibility, and cross-domain and cross-language applications. The discourse is mostly English-centred, with cross-language exceptions in German, Spanish, Portuguese, Chinese, and Japanese, largely, however, in the area of AI detection and stylometric;  \cite{irrgang2024,deSilva2024,spanish-lex-2024,qwen-lora-2025,zaitsu2023,schaaff2024,survey-ling-aitext-2025}. Second, they often rely on data for which the production process is unclear, and it is then \textit{assumed} that sudden changes in writing are due to AI. Concretely, many papers analyse scientific abstracts before and after 2022, note the stark rise of items like ``delve'', and attribute the rise to AI. This is a plausible inference, but it remains an assumption.

\subsection{Alignment, Misalignment, and Possible Influences on Humans}
The general goal of alignment is to make systems reflect human goals, values, beliefs, and broader expectations \cite{russell2015research,gabriel2020artificial}. Alignment research often proceeds through a sequence of diagnostics, characterisation, attribution, and mitigation. In digital assistants, alignment is commonly operationalised via preference learning (RLHF/DPO) \cite{ouyang2022training,rafailov2024direct}; recent work operationalises value alignment measurements for LLMs \cite{norhashim2024measuring}. However, the literature documents wider misalignments and societal harms \cite{blodgett2020language,bender2021dangers}, as well as more specific issues, such as gender and race biases in LLM behaviour and downstream domains \cite{kotek2023gender,omiye2023large}, and evaluation artefacts that confound competence with stylistic preferences \cite{perez2023discovering}.

A more benign form of misalignment is sycophancy, where AI assistants compliment users irrespective of truth; there are efforts to identify and mitigate sycophancy \cite{sharma2023sycophancy,wei2025synthetic}; similarly for verbosity \cite{park2024disentangling}. High-stakes value-level misalignment is exemplified by DeepSeek-R1 \cite{nature2025deepseek_thrills,csis2025deep_dive}: on topics central to liberal-democratic discourse, responses were reported to be censored or heavily reframed towards narratives aligned with political entities \cite{wired2025deepseek_censorship,guardian2025deepseek_tiananmen,naseh2025r1dacted,techcrunch2025deepseek_censorship}. Such model behaviour is at odds with widely endorsed free-expression norms in liberal-democratic contexts \cite{pew2025free_expression}. The societal risk is an undesired normative drift: repeated, low-salience exposure could normalise such narratives (mere-exposure/illusory-truth effects; \citealp{zajonc1968mereexposure,hasher1977frequency}), and LLM-authored political messaging might shift attitudes \cite{bai2025persuasion}.

Our work is concerned with lexical choice alignment, and in particular the steps of diagnostics, characterisation, and attribution, with pointers to mitigation. The overarching theme is that of alignment side effects, and our work is informed by the discourse around value and content alignment. Thus, structural findings on lexical (mis)alignment, particularly concerning the role of preference learning, could connect to issues of value and content (mis)alignment and could be of value for those discourses.

\subsection{Relation to AI detection}
In addition, a large literature explores AI detection (classifier-based, watermarking, perplexity/stylometry) and a large body of domain-specific corpora for detection exists (inter alia, \citealp{gehrmann-etal-2019-gltr,chakraborty2023possibilities,pmlr-v202-mitchell23a,kirchenbauer_reliability_2023,huang-etal-2025-magret}). Strengths of AI detection include available automation and analysis of multiple linguistic and extra-linguistic levels; limitations include high false-positive rates and relatively low robustness across domains and languages \cite{sadasivan2023can,weber2023testing}. Detection work addresses attribution, not linguistic-level-specific behaviour and stage-specific mechanisms. While both detection and diagnostics/identification approaches, such as our work, share the goal of automated, scalable pipelines, their overarching aims differ (detection:\ classification of instances; diagnostics/identification:\ identifying macro-level characteristics of model behaviour).

\begin{table*}[t]
\centering
\small
\begin{tabular}{l l l l l}
\hline
\textbf{Family} & \textbf{Params} & \textbf{Type} & \textbf{Repository} & \textbf{Revision (sha7, date)} \\
\hline
Falcon-3 & 7B & Base     & \texttt{tiiuae/Falcon3-7B-Base}              & \texttt{bf3d7ed} 24-12-17 \\
         &    & Instruct & \texttt{tiiuae/Falcon3-7B-Instruct}          & \texttt{1e57a0e} 25-05-31 \\
Gemma-3  & 4B & Base     & \texttt{google/gemma-3-4b-pt}                & \texttt{cc012e0} 25-03-21 \\
         &    & Instruct & \texttt{google/gemma-3-4b-it}                & \texttt{093f9f3} 25-03-21 \\
Llama-3.1 & 8B & Base    & \texttt{meta-llama/Llama-3.1-8B}             & \texttt{d04e592} 24-10-16 \\
          &    & Instruct& \texttt{meta-llama/Llama-3.1-8B-Instruct}    & \texttt{0e9e39f} 24-09-25 \\
Mistral  & 7B & Base     & \texttt{mistralai/Mistral-7B-v0.3}           & \texttt{caa1feb} 25-07-24 \\
         &    & Instruct & \texttt{mistralai/Mistral-7B-Instruct-v0.3}  & \texttt{0d4b76e} 25-07-24 \\
OLMo-2   & 7B & Base     & \texttt{allenai/OLMo-2-1124-7B}              & \texttt{7df9a82} 25-01-06 \\
         &    & Instruct & \texttt{allenai/OLMo-2-1124-7B-Instruct}     & \texttt{470b1fb} 25-01-06 \\
Yi-1.5   & 6B & Base     & \texttt{01-ai/Yi-1.5-6B}                     & \texttt{157a3d7} 24-06-26 \\
         &    & Instruct & \texttt{01-ai/Yi-1.5-6B-Chat}                & \texttt{771924d} 24-08-27 \\
\hline
\end{tabular}
\caption{An overview of the models used, with model size in parameters, model types, repositories, and pinned revisions. Short SHAs (last 7 characters) and release dates are given.}
\label{tab:models}
\end{table*}

\section{Task and Data}
\label{sec:task-and-data}

We address the gaps identified in Section~\ref{sec:related} by introducing curation-free, assumption-light metrics that cover both the \emph{what} and \emph{why} of lexical (mis)alignment. Our data comprise PubMed abstracts, which enables a comparison with the literature on lexical overuse in LLMs, which largely focuses on this very domain (Scientific English). We sample 42{,}000 abstracts from 2012-2021, i.e., the ten years preceding the release of ChatGPT. Per year, 4,200 abstracts were, without replacement, sampled. Each model (as per Table~\ref{tab:models}) generated the second part of an abstract, totalling a data basis of 63.4 million lemmatised tokens. Due to the windowing approach described in Section~\ref{sec:eval}, with a window size of 50 lemmatised tokens, this set-up provides coverage of approximately 2m lemmatised tokens per model variant after accounting for exclusions. Each abstract was split at the sentence boundary closest to its midpoint using a Python script. The first halves served as prompts for model generation; the second halves constitute the human gold standard. This design is related to cloze-style and human-continuation evaluation, where model predictions are assessed against paired human responses \citep{eisape2020cloze,giulianelli2023comes,ilia2024predict}. Because sentence boundary detection is not fully error-free, we manually spot-checked 100 abstracts. In 96 cases, the split was at a true sentence boundary; in 4 cases, the splits were non-ideal. We treat split variance as a minor source of preprocessing error, unlikely to materially affect the downstream results.

\subsection{Models and Decoding Policy}
For each prompt index \(r\in\{1,\dots,R\}\), we generate one continuation per model. The basic computational set-up is described in Section~\ref{sec:supp}. We used deterministic greedy decoding throughout. Sampling was disabled (temperature fixed at \(0\)), and nucleus and top-\(k\) controls were turned off. Decoding terminated on the model's end-of-sequence token (<eos>) or at the model token limit \(T\). Through <eos> suppression, we enforced a minimum of 120 tokens, and we capped outputs at 200 tokens. To avoid generation loops, we enforced a 4-gram no-repeat constraint. Variation was further minimised by fixing a global seed. The pad token was set equal to <eos>. These settings are consistent across all generations.

Base models are next-token generators and simply received the prompt \(r_i\) as input for plain next-token continuation. For instruct models, we used a minimal wrapper:\ a system message (``Reply only with the continuation; do not repeat the user text; no preface.'') and a user message containing the first half of an abstract \(r_i\) were passed. Loop/meta removal, safety messages, and meta-chat were handled symmetrically during the cleaning stage.

\subsection{Model Families} 

Model families were selected to (i) provide both base and instruct variants and (ii) support temperature \(=0\) decoding to maximise reproducibility. Under these criteria, we chose six popular families:\ Falcon \cite{almazrouei2023falcon}, Gemma \cite{gemma2025techreport}, Llama-3.1 \cite{grattafiori2024llama3}, Mistral-v0.3 \cite{jiang2023mistral7b}, OLMo-2 \cite{teamolmo2025olmo2}, and Yi-1.5 \cite{young2024yi}. We used the mid-sized models; in these families, larger variants typically add multimodality with limited improvement on purely textual tasks \cite{huggingface_open_llm_leaderboard_2024}. This gives the base-instruct pairs listed in Table~\ref{tab:models}.

\subsection{Cleaning and Part-of-Speech Tagging}
\label{sec:cleaning}

Cleaning was deletion-only and symmetric (to avoid measuring pre-processing artefacts), i.e., applied to the human gold standard as well as to both base and instruct generations. Cleaning proceeded in two stages. First, we applied a deterministic regex pre-clean:\ normalise whitespace; collapse runs and newlines; strip leading and trailing spaces; and drop any text following the most frequent signal that the abstract has concluded and the model is continuing into the article body (``Introduction''). All removals were logged. Second, for information that is difficult to remove reliably with pattern matching, we used GPT-4.1-mini with temperature set to $0$ to delete AI persona and meta text (``Certainly, here is \dots''), dialogue scaffolding (``\textless assistant\textgreater''), loops (keeping one copy), and first- and second-person material (``Can you explain the meaning of \dots''). We instructed the cleaner to preserve genuine abstract content and not to paraphrase or reorder (outputs include diffs and run summaries). The exact GPT-4.1 system prompt follows best practices and can be found in Appendix~A. The cleaning prompt was simply:

\begin{lstlisting}
"Clean the following MID-ABSTRACT CONTINUATION by applying ONLY the deletion rules."
"Do not paraphrase or rewrite; return the cleaned text only.\n\n"
f"INPUT:\n{raw}\n\n"
"OUTPUT (cleaned text only):"
\end{lstlisting} 

\noindent To analyse different inflected forms (e.g., `delve', `delved', `delving') under a single lemma type, and to distinguish words that share a surface form but differ in meaning/usage (e.g., `analysed' as a verb vs an adjective), we tag the cleaned generations for part-of-speech (POS). This also enables analyses of additional categories (e.g., excluding certain part-of-speech categories from certain computational steps). We used spaCy 3.8 (\texttt{en\_core\_web\_trf}; \citealp{honnibal2020spacy}) for tagging, with the Universal Part-of-Speech (UPOS) tags from the Universal Dependencies framework \cite{nivre-etal-2020-universal}; outputs are in CoNLL\textendash U format \cite{zeman-etal-2017-conll}.

\section{Evaluation Metrics}
\label{sec:eval}

Both metrics introduced below, the \emph{Lexical Alignment Score} (LAS) and the \emph{Triangulated Preference Shift} (TPS), operate on length-controlled windows of paired human-model continuations, using windowed document prevalence to ensure robustness against few-document spikes. 

Intuitively, the two metrics approach the following questions. LAS asks whether a model uses a given lexical item more or less often than humans do under the same prompt. TPS asks whether such overuse appears to arise mainly after the base stage, rather than already being present in the base model. In this sense, LAS targets the \emph{what} of lexical divergence, whereas TPS targets part of the \emph{why}.

\subsection{Windowed Document Prevalence}
\label{sec:windows}

Per abstract, the first half serves as the prompt; the second half is the human gold standard $H$. Using the prompt, we generate continuations for the base model ($B$) and instruct model ($I$). 

Plain relative frequencies are fragile:\ as corpus size grows, the probability of distortions rises whenever a plausible term occurs repeatedly in only one \textit{stream} $S$ (human/base/instruct); say e.g., two prompts elicit multiple ``EuroQol'' mentions in $I$ but not in $H$ or $B$. A reasonable fix for this is \emph{document frequency} (did the item occur at all?); however, models may systematically produce documents of different lengths, giving items varying degrees of opportunity to occur. We therefore score presence within a fixed-size window, $K{=}50$ lemmatised tokens by default (the code allows variation), placed at a random region of the document (with a deterministic seed for the purposes of this work). Concretely, we choose a percentile offset $\pi_r\in(0,1)$ deterministically for the prompt ID and use the same $\pi_r$ across $H$-$B$-$I$ for that prompt, in order to ensure paired/triplet comparability (pairs for LAS; triplets for TPS). Pre-processing is symmetric across $H/B/I$ (see Section~ref{sec:cleaning}); if any continuation is too short after cleaning, the entire pair/triplet is removed. This gives a length-controlled, spike-robust prevalence signal. An informal example of the window approach (using words instead of lemmatised tokens) is as follows. Suppose the entire document is ``This is a sentence with seven words'', and the window size is $K{=}3$. The sentence has 7 words, so there are $7 {-} 3 {+} 1 = 5$ possible 3-word windows. A random number between 0 and 1 is drawn, say $0.70$ (the 70th percentile). Multiplying this by $5$ and rounding down gives a starting index of $3$ (counting from $0$). This corresponds to positions $4-6$ when counting from the first word. Taking these three words gives the window:\ ``This is a [sentence with seven] words''.

\vspace{-0.3cm}

\paragraph{Formal Definition.} We write $S\in\{H,B,I\}$ for the stream and index prompts by $r=1,\dots,R$. Each pair $(r,S)$ constitutes a document $D_{r,S}$ (the human second half if $S{=}H$, or the model continuation if $S\in\{B,I\}$). A stream $S$ is the collection $\{D_{r,S}:\ r=1,\dots,R\}$. Let $w$ denote a lemma type with UPOS tag, and let \(\omega(\cdot)\) be the lemmatisation map from lemmatised tokens to lemma types; for a token \(t\), \(\omega(t)\) is its lemma+UPOS type. For prompt $r\in\{1,\dots,R\}$ and stream $S$, we define:\
\[
\begin{aligned}
Y_{rS}(w) &= \mathbf{1}\!\Bigl\{\text{$w$ appears in the window of size}\\
          &\quad \text{$K$-lemmatised tokens of $S$ at perc.\ $\pi_r$}\Bigr\}\\
c_S(w)    &= \sum_{r=1}^{R} Y_{rS}(w)
\end{aligned}
\]
Intuitively, $Y_{rS}(w)$ is a per-document window flag:\ $1$ if $w$ occurs at least once in the $K$-lemmatised-token window of document $D_{r,S}$, else $0$. Summing over prompts gives $c_S(w)=\sum_r Y_{rS}(w)$, the windowed document frequency for stream $S$. For example, if ``EuroQol'' appears many times in the $I$-window for two prompts only, then $c_I(\text{EuroQol})=2$. Here $R$ is the number of retained prompts after symmetric cleaning, so $c_S(w)\in\{0,1,\dots,R\}$. To avoid degenerate $0/1$ rates (infinite log-odds) and reduce small-sample variance for rare lemmas, we estimate window-prevalence with Jeffreys smoothing (a common smoothing method for document-frequency estimation, \citealp{krichevsky1981}), which is essentially adding a half ``pseudo-hit'' and a half ``pseudo-miss'' to stabilise the estimate, with Beta$(\tfrac12,\tfrac12)$ prior:
\[
\ell_S(w)=\frac{c_S(w)+\tfrac{1}{2}}{R+1}
\]

\noindent where $\ell_S(w)$ is the Jeffreys-smoothed \emph{windowed prevalence} of $w$ in stream $S$. 

\subsection{Two Stages}
\label{sec:two-stages}

Both LAS and TPS run in two consecutive stages.

\noindent \textbf{Estimation.} On the cleaned $H/B/I$ documents, we compute $\ell_S(w)$ for \emph{all} UPOS-tagged lemma-types (incl.\ \textsc{PUNCT}). UPOS usage (e.g., punctuation rates) may differ between human and model texts; filtering here could induce selection effects and artefactual convergence/divergence even for non-filtered categories. This gives us windowed document-prevalence estimates for each stream-lemma pair $(S,w)$, which are the basis for both LAS and TPS. Programmatically, this step builds a look-up table from lemma-types (lemma+UPOS) to (i) $\ell_S(w)$ and to (ii) derived per-lemma-type contrasts used by the two metrics, LAS and TPS. 

\noindent \textbf{Scoring.} For both LAS and TPS, we use the estimated $\ell_S(w)$ to score lemmatised tokens, sequences, documents, and corpora/models; essentially, using the look-up table for scoring. We report on the full tag set; the code allows tag selections.

\begin{table}
\centering
\begin{minipage}{\columnwidth}\centering
  \subcaptionbox{Top-20 wLAS entries, aggregated over all instruct models.\label{tab:wlasinstruct}}{%
    \small
    \setlength{\tabcolsep}{6pt}
    \begin{tabular}{r l r}
    \hline
    \hline
    \textbf{Rank} & \textbf{lemma+UPOS} & \textbf{wLAS} \\
    \hline
    1  & these\_DET        & 0.162 \\
    2  & to\_PART          & 0.123 \\
    3  & suggest\_VERB     & 0.093 \\
    4  & finding\_NOUN     & 0.080 \\
    5  & this\_DET         & 0.078 \\
    6  & the\_DET          & 0.078 \\
    7  & a\_DET            & 0.077 \\
    8  & furthermore\_ADV  & 0.077 \\
    9  & such\_ADJ         & 0.076 \\
    10 & research\_NOUN    & 0.076 \\
    11 & additionally\_ADV & 0.074 \\
    12 & to\_ADP           & 0.070 \\
    13 & for\_ADP          & 0.070 \\
    14 & further\_ADJ      & 0.069 \\
    15 & study\_NOUN       & 0.069 \\
    16 & that\_SCONJ       & 0.065 \\
    17 & ,\_PUNCT          & 0.061 \\
    18 & could\_AUX        & 0.061 \\
    19 & into\_ADP         & 0.060 \\
    20 & highlight\_VERB   & 0.059 \\
    \end{tabular}
  }

\vspace{0.5em}

  \subcaptionbox{Top-20 wLAS entries for nouns, verbs, adjectives, adverbs, aggregated over instruct models.\label{tab:wlas2}}{%
    \small
    \setlength{\tabcolsep}{6pt}
    \begin{tabular}{r l r}
    \hline
    \hline
    \textbf{Rank} & \textbf{lemma+UPOS} & \textbf{wLAS} \\
    \hline
 3  & suggest\_VERB        & 0.093 \\
 4  & finding\_NOUN        & 0.080 \\
 8  & furthermore\_ADV     & 0.077 \\
 9  & such\_ADJ            & 0.076 \\
10  & research\_NOUN       & 0.076 \\
11  & additionally\_ADV    & 0.074 \\
14  & further\_ADJ         & 0.069 \\
15  & study\_NOUN          & 0.069 \\
20  & highlight\_VERB      & 0.059 \\
21  & potential\_ADJ       & 0.055 \\
22  & include\_VERB        & 0.053 \\
24  & lead\_VERB           & 0.053 \\
25  & crucial\_ADJ         & 0.053 \\
28  & understand\_VERB     & 0.049 \\
30  & role\_NOUN           & 0.048 \\
31  & enhance\_VERB        & 0.047 \\
32  & involve\_VERB        & 0.045 \\
34  & researcher\_NOUN     & 0.045 \\
35  & various\_ADJ         & 0.044 \\
36  & strategy\_NOUN       & 0.044 \\
    \end{tabular}
  }
\end{minipage}
\caption{Top lexical shifts by lemma-type (wLAS). Higher values indicate model overuse vs human.}
\label{tab:wlas}
\end{table}

\subsection{Lexical Alignment Score}
\label{sec:las}

The Lexical Alignment Score quantifies the alignment of a model $M$ (either $B$ or $I$) relative to $H$ (while one could compare $B$ and $I$ directly, $H$ is the natural reference; for three-way contrasts, the Triangulated Preference Shift metric is more useful). Estimation is performed at the lemma+UPOS level and later used for scoring lemmatised tokens, sequences, documents, and corpora/models.

\vspace{-0.3cm}

\paragraph{Estimation.}
Estimation is per UPOS-tagged lemma-type, using the paired prompt-aligned windows defined in Section~\ref{sec:windows}. For a lemma type $w$ and a model $M\in\{B,I\}$ we define the lemma-type-level LAS:
\[
\mathrm{wLAS}_M(w)\;=\;\ell_M(w)\;-\;\ell_H(w)
\]
where $\ell_S(w)$ is the Jeffreys-smoothed windowed prevalence in stream $S$. Positive values indicate overuse relative to human usage; negative values indicate underuse. The following example illustrates this:\ Suppose that a lemma occurs in 18\% of model windows but in 10\% of human windows, and is thus overused. Then its word-level LAS is \(0.18 - 0.10 = 0.08\). If the values were instead 7\% versus 10\%, thus underuse, the LAS score would be \(-0.03\).

\vspace{-0.3cm}

\paragraph{Scoring.}
Given these per-lemma-type scores $\mathrm{wLAS}_M(w)$, we define per-lemmatised-token contributions
$\Delta(t){=}\mathrm{wLAS}_M\!\big(\omega(t)\big)$ for lemmatised token $t$ in a unit $U$
(lemmatised tokens, sequence, document, or corpus) with lemmatised-token multiset $\mathcal{T}(U)$. We score $U$ by the
\emph{L2 mean} \cite{bishop2006pattern}: 
\[
\mathrm{uLAS}(U;M)
= \left( \frac{1}{\lvert \mathcal{T}(U)\rvert}
    \sum_{t \in \mathcal{T}(U)} \Delta(t)^{2}
  \right)^{\!1/2}
\]

\noindent This gives a length-normalised root-mean-square (RMS) magnitude that avoids cancellation between over- and under-use, while placing greater emphasis on large deviations. Aliases for readability:\ \(\mathrm{sLAS}(s;M)=\mathrm{uLAS}(s;M)\) (sequence),
\(\mathrm{dLAS}(d;M)=\mathrm{uLAS}(d;M)\) (document),
and \(\mathrm{cLAS}(C;M)=\mathrm{uLAS}(C;M)\) (corpus). 

\vspace{-0.3cm}

\paragraph{Model Aggregation.}
Scores may be aggregated across models; we take a simple macro-average, giving equal weight to every model irrespective of the number of documents/lemmatised tokens analysed. This is most informative at the lemma type level, which gives an overview of the most misaligned lemmas across LLMs. Section~\ref{sec:results} reports model-aggregated \(\mathrm{wLAS}\) and the strongest shifts.

\vspace{-0.3cm}

\paragraph{Scoring Example.}
$s_1$ and $s_2$ illustrate sequence-level scoring (casing/lemmatisation/cleaning omitted).
Numbers in the second row are the per-lemmatised-token contributions $\mathrm{wLAS}_M(\omega(t))$.
$\mathrm{uLAS}$ is the length-normalised L2 mean.
\vspace{-0.1cm}
\[
\begin{array}{lccccc}
s_1: & \texttt{Our} & \texttt{work} & \texttt{shows} & \texttt{how} & \cdots \\
\Delta(t)     & -0.027        & -0.005        & -0.002         & 0.001 & \cdots
\end{array}
\]
\vspace{-0.3cm}
\[
\begin{array}{lccccc}
s_2: & \texttt{Our} & \texttt{work} & \texttt{highlights} & \texttt{how} \\
\Delta(t)     & -0.027        & -0.005        & 0.027         & 0.001
\end{array}
\]

\noindent We score each sequence with the \(L2\) mean:\ square \(\Delta(t)\) values, average over the lemmatised token count (\(\lvert\mathcal{T}(s_i)\rvert=4\)), then take the square root, resulting in \(\mathrm{sLAS}(s_1)\approx 0.014\) and \(\mathrm{sLAS}(s_2)\approx 0.019\).

\subsection{Triangulated Preference Shift}
\label{sec:tps}

The Triangulated Preference Shift metric isolates preference-stage effects by requiring an instruction-tuned model $I$ to exceed both the human baseline $H$ and the base model $B$, while penalising shifts already present in $B$. Estimation is performed at the lemma+UPOS level and then used for scoring different linguistic units.

\vspace{-0.3cm}

\paragraph{Estimation.}
Estimation is done for lemma-types $w$ and a model family $M$ with streams $B_M$ (base), $I_M$ (instruct), and a common human baseline $H$. With $\ell_S(w)$ being the Jeffreys-smoothed windowed prevalence in stream $S$ (Section~\ref{sec:windows}), we get:\ 

\[
\begin{aligned}
\Delta_{IH}(w)=\ell_{I}(w)-\ell_H(w) \\
\Delta_{IB}(w)=\ell_{I}(w)-\ell_{B}(w) \\
\Delta_{BH}(w)=\ell_{B}(w)-\ell_H(w)
\end{aligned}
\]

\noindent The lemma-type-level triangulated pref.\ shift is:\

\vspace{-0.2cm}

\[
\begin{alignedat}{2}
\mathrm{wTPS}_{M}(w)\; &={} & \min\!\big\{\Delta^{(M)}_{I\!H}(w),\,\Delta^{(M)}_{I\!B}(w)\big\} \\
                       &     & {}-\max\!\big\{0,\,\Delta^{(M)}_{B\!H}(w)\big\}
\end{alignedat}
\]

\begin{table}[htbp]
\centering
\small
\begin{tabular}{r l r}
\hline
\hline
\textbf{Rank} & \textbf{lemma+UPOS} & \textbf{wTPS} \\
\hline
1 & to\_PART & 0.130 \\
2 & these\_DET & 0.112 \\
3 & this\_DET & 0.085 \\
4 & to\_ADP & 0.076 \\
5 & furthermore\_ADV & 0.075 \\
6 & such\_ADJ & 0.073 \\
7 & research\_NOUN & 0.070 \\
8 & for\_ADP & 0.069 \\
9 & additionally\_ADV & 0.065 \\
10 & further\_ADJ & 0.064 \\
11 & study\_NOUN & 0.061 \\
12 & ,\_PUNCT & 0.058 \\
13 & highlight\_VERB & 0.057 \\
14 & into\_ADP & 0.057 \\
15 & potential\_ADJ & 0.056 \\
16 & finding\_NOUN & 0.055 \\
17 & a\_DET & 0.054 \\
18 & as\_ADP & 0.054 \\
19 & could\_AUX & 0.053 \\
20 & crucial\_ADJ & 0.052 \\
\end{tabular}
\caption{Top 20 overused lemma+UPOS by wTPS.}
\label{tab:top20-wtps-simple}
\end{table}

\noindent Informal reading:\ wTPS measures the portion of the instruct model's uplift that cannot be explained by the base model. It is positive only when $I$ beats both $H$ and $B$, and by more than any pre-existing $B>H$ lean; otherwise it is zero or negative.

The following example illustrates the score. Suppose that a lemma occurs in 10\% of human windows, 12\% of base-model windows, and 20\% of instruct-model windows. Then \(\Delta_{IH}=0.10\), \(\Delta_{IB}=0.08\), and \(\Delta_{BH}=0.02\). TPS is therefore \(\min(0.10,0.08)-0.02 = 0.06\), i.e.\ a positive residual shift beyond what was already present in the base model. By contrast, if the base model were already at 18\%, TPS would be much smaller or vanish.

\vspace{-0.3cm}

\paragraph{Scoring.} Given $\mathrm{TPS}(w)$, we compute an $L2$ root-mean-square (RMS) score $\mathrm{uTPS}$ that applies uniformly to sequences, documents, and corpora (similar to $\mathrm{uLAS}$). For any unit $U$ with lemmatised token multiset $\mathcal{T}(U)$ and lemmatisation map $\omega(\cdot)$:  

\vspace{-0.2cm}

\[
\mathrm{uTPS}(U; M)
=\left(\frac{1}{|\mathcal{T}(U)|}\sum_{t\in\mathcal{T}(U)}
\mathrm{wTPS}_{M}\!\big(\omega(t)\big)^{2}\right)^{\!1/2}
\]

\noindent This gives a length-normalised RMS magnitude that avoids cancellation between over- and under-use, and weights larger deviations more strongly. For reporting, we focus on upward shifts, using $\max(0,\mathrm{wTPS}(w))$. Aliases:\ $\mathrm{sTPS}(s)=\mathrm{uTPS}(s)$ (sequence), $\mathrm{dTPS}(d)=\mathrm{uTPS}(d)$ (document), and $\mathrm{cTPS}(C)=\mathrm{uTPS}(C)$ (corpus).

\vspace{-0.3cm}

\paragraph{Model Aggregation.}
As with LAS, TPS scores may be aggregated across models.

\section{Results}
\label{sec:results}
\begin{figure*}[t]
  \centering
  \includegraphics[width=0.8\textwidth]{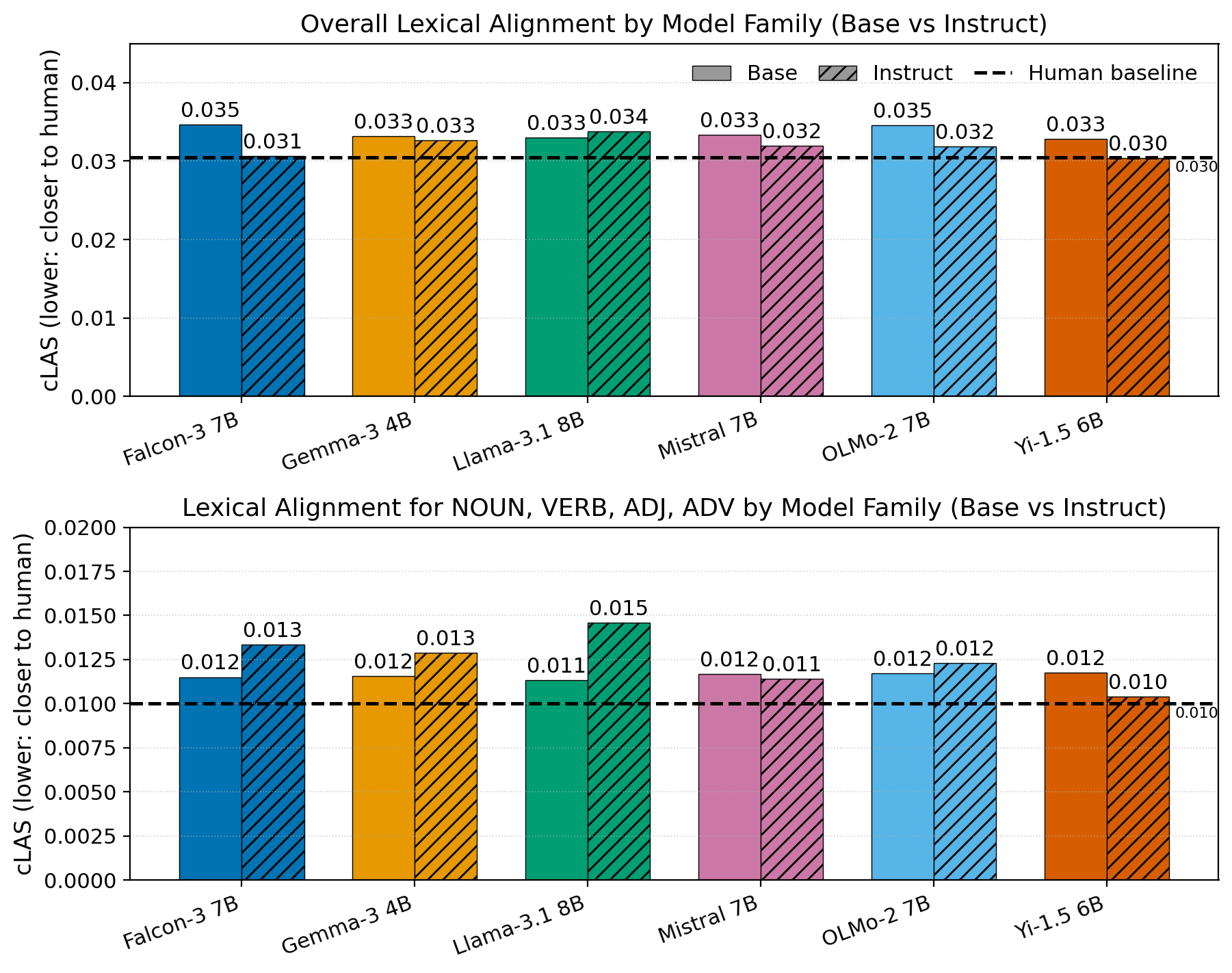}
  \caption{cLAS by model family (base vs instruct); lower:\ closer to human.}
  \label{fig:clas_barplot}
\end{figure*}

\begin{figure}[t]
  \centering
  \includegraphics[width=\columnwidth]{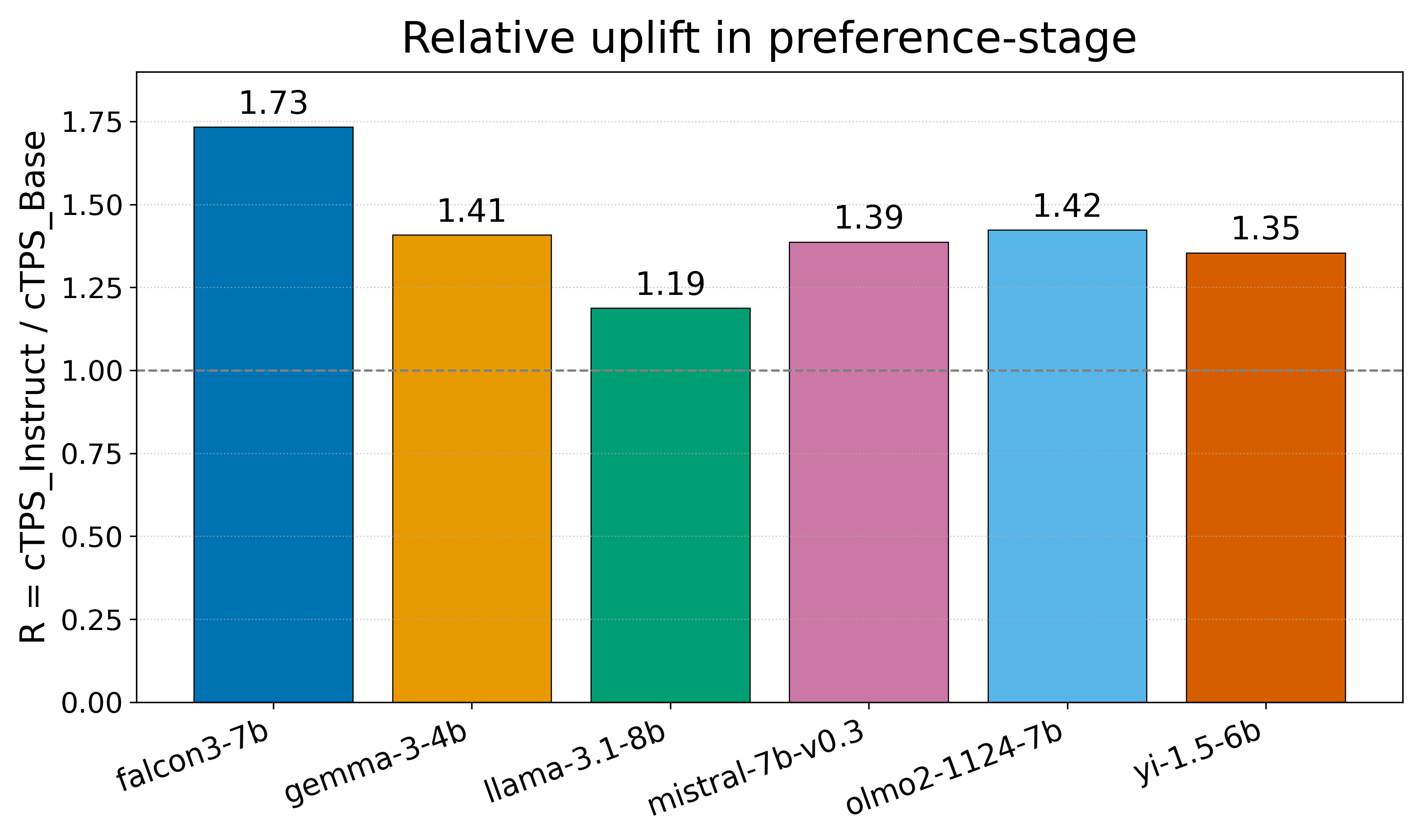}
  \caption{cTPS ratios (base vs instruct).}
  \label{fig:clas_barplot3}
\end{figure}

Table~\ref{tab:wlasinstruct} reports the top 20 lemma-level Lexical Alignment Scores aggregated over all instruct models; results for the base models can be found in Table~\ref{tab:wlasbase} in Appendix~B. Table~\ref{tab:wlas2} links to prior work by limiting analysis to content words (\textsc{NOUN}/\textsc{VERB}/\textsc{ADJ}/\textsc{ADV}) and aggregating over instruct models. The top 20 items ranked by the preference-stage metric are listed in Table~\ref{tab:top20-wtps-simple}. Macro-level alignment, corpus-level LAS, by family and stage is shown in Figure~\ref{fig:clas_barplot} (lower values indicate closer alignment to humans). Figure~\ref{fig:clas_barplot3} shows the corpus-level TPS \textit{ratio} $R$, defined as $R=\mathrm{cTPS}_{\text{instruct}}/\mathrm{cTPS}_{\text{base}}$, where higher values of $R$ indicate stronger preference-stage shifts in the instruct variant. The results show that the proposed diagnostics identify lexical items already discussed in the literature and that, although instruct models are often closer to humans overall, this pattern weakens or reverses when attention is restricted to content words. 

\section{Validation}
\label{sec:validation}

To further validate the metrics, we conducted four checks. (i) we scored an additional, unseen 20\% of the data. (ii) We compare our overuse list to the literature. (iii) We ran the metrics with different window sizes ($K=40, 50, 60$). (iv) We reran the $K=50$ configuration for four more random seeds. The following is done for the LAS metric.

\vspace{-0.3cm}

\paragraph{Additional Data.}
An additional 20\% of data, comprising 8{,}400 abstracts (840 per year), was scored. The results per model are (base vs instruct):\ Falcon:\ 0.0346 vs 0.0306, Gemma:\ 0.0332 vs 0.0326, Llama:\ 0.0331 vs 0.0338, Mistral:\ 0.0333 vs 0.0320, OLMo:\ 0.0345 vs 0.0317, Yi:\ 0.0329 vs 0.0304. The human baseline comes out at 0.0304. The results are extremely similar to those reported in Figure~\ref{fig:clas_barplot}. 

\vspace{-0.3cm}

\paragraph{Convergence with Prior Research.}
Prior work provides curated inventories of AI-associated words. \citet{geng2024chatgpt} discuss 8 words in-depth. \citet{galpin2025exploring} list 32 overused words (surface-form level, i.e., including inflections), \citet{kobak2024delving} list 291 such items. We matched these against our overused set (focus: instruct models); for ambiguous surface forms (e.g., ‘diminishing’, which can occur as a verb or an adjective), we counted a hit if any corresponding lemma+UPOS variant appeared. 8/8 words from \citet{geng2024chatgpt} feature prominently in our data. From the list in \citet{galpin2025exploring}, 32/32 entries occur in our list. From the 291 items in \citet{kobak2024delving}, 240 (83\%) occur in our data. 

\vspace{-0.3cm}

\paragraph{Varying $K$.}
We observe that scores vary slightly across different window sizes. Here, we choose $K=40$ (smaller windows disregard a larger portion of the data), $50$, and $60$ (window sizes larger than that approach document length and undermine the windowing logic). The scores averaged across model families for base vs instruct models (with the human baseline in parentheses) are:\ 0.357 vs 0.354 (0.324), 0.336 vs 0.319 (0.3048), and 0.324 vs 0.318 (0.305), respectively. The general trend that base > instruct > human remains consistent across window sizes. Some variation is expected, partly due to the changing proportional effect of Jeffreys smoothing as window size increases. 

\vspace{-0.3cm}

\paragraph{Different Seeds.}
The global average over all base and instruct models at seed 42 is 0.03275. For four additional seeds, the results are 0.03276 (s=43), 0.03277 (s=44), 0.03283 (s=45), and 0.03281 (s=46). The outcomes are very robust.

\vspace{-0.3cm}

\paragraph{Further Validation.}
The TPS could be further validated by analysing instruction and preference datasets to assess whether the metric aligns with the linguistic patterns those datasets encode. The datasets used in OLMo could be particularly well suited for this purpose, as Allen AI have made the entire process, including all data, publicly available.

Additional validation could also come from controlled model development:\ deliberately inducing controlled doses of a model bias and testing whether it manifests in model behaviour. This would parallel \citet{zhang2024lists}, who analysed style- and format-related model behaviours.

\section{Discussion}
\label{sec:discussion}

At the corpus level, instruct variants generally align \emph{better} with human usage than their base counterparts (5/6 families; Figure~\ref{fig:clas_barplot} top). However, focusing on content words (\textsc{NOUN}/\textsc{VERB}/\textsc{ADJ}/\textsc{ADV}), for all models this trend either reverses (4/6 families in Figure~\ref{fig:clas_barplot} bottom) or at least weakens. These are precisely the items emphasised in prior reports of lexical overuse. However, the largest aggregate shifts are dominated by function markers (Table~\ref{tab:wlasinstruct}), consistent with the findings that AI models have their own syntax style \cite{zamaraeva2025comparing}. The Triangulated Preference Shift metric corroborates stage influence:\ many high-LAS items also show a positive TPS, indicating uplift in instruct over both human and base (Table~\ref{tab:top20-wtps-simple}), which affects some model families more than others (Figure~\ref{fig:clas_barplot3}). Per-model results are feasible and important for model-individual audits, and per-model results can be found in Appendix~C.

An important observation is that quite a few of the top-ranked items are not unusual scientific terms. We interpret this as evidence that the models amplify already typical scientific wording, pushing high-prevalence items toward even greater dominance. This observation also highlights an aspect of the metrics:\ they are most sensitive to prevalence \textit{volume} and, arguably, though this remains to be explored, less sensitive to low-frequency words that show small absolute but large relative shifts (``spikes''). Such effects might be captured more directly by ratio-based measures, for example using log prevalence ratios between human and model outputs. At the same time, even these lower-volume but higher-relative shifts may reflect a related tendency, namely the amplification of lexical patterns that are already present, at least in emerging form, in human writing. We regard both phenomena as important, and therefore emphasise the value of systematic comparisons across human texts, base models, and instruction-tuned models.

Our work is situated within the discourse on lexical overuse. In principle, the Lexical Alignment Score captures both overuse and underuse. The latter is generally less noteworthy, as it tends to show less disruptive shifts \cite{galpin2025exploring}. By contrast, the Triangulated Preference Shift metric measures foremost positive shifts that are attributable to preference learning.

A broader relevance of our work stems from the observation that AI-associated lexicon is now attested in spontaneous speech \cite{yakura2024empirical,anderson2025model}, indicating rising prevalence in human language use. While our work does not establish adoption directly, our findings identify model-side lexical biases that are plausible candidates for such uptake. AI chatbots may be accelerating, and possibly causing, these shifts; therefore, the ability to align models with human linguistic expectations is valuable. One plausible pathway is repeated exposure during AI-assisted writing and interaction:\ psycholinguistic work shows that speakers tend to align with recently encountered linguistic forms, including those produced by computer and robot interlocutors \citep{brennan1991conversation, branigan2003syntactic, brandstetter2017persistent, ostrand2023rapid}. Work on AI-mediated communication further suggests that such exposure can reduce diversity in co-writing \citep{padmakumar2023does} and may, in some settings, leave persistent linguistic traces in subsequent human communication \citep{riedl2024socialforcefield}. More broadly, our work contributes to diagnostics, characterisation, and partial attribution, mirroring core stages of alignment research in other domains. Although we do not study value alignment directly, the present findings help illuminate how preference-based post-training can generate measurable behavioural side effects, and may therefore inform wider mechanism-focused discussions of model alignment. 

A question that lies beyond the scope of this paper, but is important for the field more broadly, is whether the underlying phenomenon truly constitutes \textit{mis}alignment. The alignment procedures themselves may be effective:\ models might indeed align to the preferences present in the preference, learning datasets; preferences expressed by data workers, but these may differ from the expectations of end users. For a discussion of this issue, see \citet{santurkar2023whose} and \citet{he2024whose}.

\section{Conclusion}
\label{sec:conclusion}

The goal of this paper was to introduce evaluation metrics for measuring lexical alignment and the contribution of preference learning. Critically, the proposed methods are curation-free and assumption-light. Key to it is a scalable design in which model continuations are compared against matched human continuations from the same source documents. 

The Lexical Alignment Score and Triangulated Preference Shift offer promising approaches for quantifying both the \emph{what} (how closely does model usage match human usage) and the \emph{why} of lexical alignment (how much divergence is attributable to the instruction/preference stage), with clear potential to inform model development. Importantly, these metrics also enable the study of lexical (mis)alignment beyond Scientific English and across languages other than English. This matters because LLM-based chat assistants are now used by millions of people worldwide, who are continually exposed to their linguistic outputs. Ensuring that these outputs align with human expectations is therefore of growing societal importance.

\section*{Acknowledgements}

We thank Gordon Erlebacher and Zina Ward for their valuable input, and the reviewers for their very constructive feedback. We are grateful to the Florida State University Research Computing Center for computational support. 

\section*{Code, Data, Computing Set-up}
\label{sec:supp}

\noindent \textbf{Code and Data.} All code, with notes on how to retrieve data, is available at:\ \href{https://github.com/fsu-nlp/lexical-alignment-shifts}{github.com/fsu-nlp/lexical-alignment-shifts}.  

\noindent \textbf{Computational Set-up.} All major computations were run on two machines.

\noindent\textbf{(A) GPU server.}
NVIDIA H100 PCIe (80\,GB); driver~570.148.08; CUDA~12.8. 
Intel Xeon Platinum 8480+; 221\,GiB RAM. 
Ubuntu~24.04.2 LTS; Linux~6.11.0-29.

\noindent\textbf{(B) University HPC node.}
NVIDIA RTX~A4500 (20\,GB). 
AMD EPYC~7313 (16 cores); 251\,GiB RAM. 
Linux~4.18.0-372.32.1.

\noindent\textbf{Software (identical on both).}
Python~3.12.3; PyTorch~2.8.0{+}cu128 (CUDA~12.8; cuDNN~91002); 
\texttt{transformers}~4.56.1; \texttt{accelerate}~1.10.1; 
\texttt{peft}~0.17.1; \texttt{spaCy}~3.8.7.

\section*{Ethical Considerations}

This work presents minimal risks:\ data are public PubMed abstracts and model outputs; procedures are open and aimed at reducing bias and improving fairness; misuse potential is low; and compute was reasonable (six models; cost about \$1160). A broader ethical concern, however, is the labour behind preference-learning datasets, which is often precarious \cite{perrigo2025sweatshop,perrigo2023kenya}. While not inherent to the technology, this is part of its current realisation and warrants continued scrutiny.

\section*{Limitations}

Our work has several limitations, of which we discuss the most notable ones. Firstly, the present design is limited in that the human gold standard always comes from the second half of the abstract. This may introduce position effects:\ some lexical items are more likely to occur early in abstracts, whereas others are more likely to occur later. For instance, markers such as \emph{firstly} may be associated with earlier text positions, while items such as \emph{lastly} may be associated with later ones. Some divergences may therefore partly reflect document-position asymmetries rather than model behaviour alone. Future work should extend the design to windows or gold-standard continuations drawn from all text positions.

Second, it is restricted to Scientific English; and more precisely, to the language of continuations of Scientific English abstracts, as abstract-initial language will most likely be scarce in our data. This restriction is motivated both by cost considerations (running six models within a single domain well exceeded \$1{,}000) and by the need to ensure comparability with the existing literature. That said, the project is expressly motivated by scaling this line of work beyond Scientific English. 

Third, we do not include the most popular chat assistant, ChatGPT, because since GPT-4, there is no base variant public, which prevents triangulation, and because since GPT-5, temperature cannot be set to 0 (it has a temperature parameter, it just cannot be set to 0), which hampers reproducibility. 

Further, there is a very high likelihood of some overlap between our PubMed-based evaluation set and model pre-training data. This may raise absolute human–model similarity, especially for base models; however, it is less likely to account for within-family base-instruction contrasts, which primarily reflect post-pre-training changes.

\section*{Appendix A: System Prompt Cleaning}
\label{app:systemprompt}

For the symmetric cleaning step, generated and human continuations were processed with the following system prompt:

\begin{quote}
\small
\ttfamily
ROLE: Editorial cleaner for SCIENTIFIC ABSTRACT CONTINUATIONS (mid-abstract).\\
ACTIONS: ONLY DELETE text; NEVER paraphrase, reorder, merge, or add. Keep sentence wording and order. If uncertain whether to delete, KEEP.\\[0.5em]

DELETE:\\
1) Meta/AI persona: e.g., ``Certainly, here is ...'', ``as an AI model'', apologies, instructions, tool/safety notes.\\
2) Conversation turns \& scaffolding (only if truly dialogic): pseudo-dialogue markers \texttt{<|user|>}, \texttt{<|assistant|>}, \texttt{</|user|>}, \texttt{</|assistant|>} ONLY IF followed by chat-like material (greeting, instruction, apology, question to reader); delete the marker and that span; otherwise delete markers only.\\
3) Obvious repetition/loops: remove verbatim or near-verbatim repeats; KEEP one copy.\\
4) First/second-person META sentences using ``I/me/my'' or direct address ``you/your''. Do NOT delete ``we/our/us''.\\[0.5em]

PRESERVE:\\
- Keep phrases like ``In conclusion'' / ``concluding'' when embedded in a normal sentence.\\
- All scientific content and phrasing (incl.\ ``we/our/us'').\\
- Angle-bracketed tokens in general (operators, tags, gene symbols, XML-like markup) EXCEPT the pseudo-dialogue markers listed above.\\
- Original wording, punctuation, and order. Do NOT fix grammar, reflow text, or change casing.\\[0.5em]

OUTPUT: cleaned text ONLY (no quotes, no notes). If nothing but commentary remains, output an empty string.
\end{quote}

\section*{Appendix B: wLAS for Base Models}
\label{app:wlasbase}

\begin{table}[!htbp]
\centering
\small
\setlength{\tabcolsep}{6pt}
\begin{tabular}{r l r}
\hline
\hline
\textbf{Rank} & \textbf{lemma+UPOS} & \textbf{wLAS} \\
\hline
    1  & result\_NOUN       & 0.132 \\
    2  & that\_SCONJ        & 0.127 \\
    3  & be\_AUX            & 0.127 \\
    4  & suggest\_VERB      & 0.096 \\
    5  & also\_ADV          & 0.094 \\
    6  & the\_DET           & 0.079 \\
    7  & than\_ADP          & 0.060 \\
    8  & significantly\_ADV & 0.048 \\
    9  & 10\_NUM            & 0.034 \\
    10 & show\_VERB         & 0.034 \\
    11 & conclusion\_NOUN   & 0.031 \\
    12 & mean\_ADJ          & 0.031 \\
    13 & may\_AUX           & 0.030 \\
    14 & high\_ADJ          & 0.027 \\
    15 & most\_ADV          & 0.026 \\
    16 & addition\_NOUN     & 0.026 \\
    17 & common\_ADJ        & 0.023 \\
    18 & .\_PUNCT           & 0.022 \\
    19 & these\_DET         & 0.022 \\
    20 & a\_DET             & 0.021 \\
\end{tabular}
\caption{Top-20 wLAS entries, aggregated over all base models. Higher values indicate greater overuse in model output relative to human continuations.}
\label{tab:wlasbase}
\end{table}

\section*{Appendix C: Model-Specific Results}
\label{app:modelspecific}

Higher values indicate greater overuse in model output relative to human continuations.

\begin{table}[!htbp]
\centering
\small
\setlength{\tabcolsep}{6pt}
\begin{tabular}{r l r}
\hline
\hline
\textbf{Rank} & \textbf{lemma+UPOS} & \textbf{wLAS} \\
\hline
1  & be\_AUX            & 0.142 \\
2  & result\_NOUN       & 0.115 \\
3  & also\_ADV          & 0.114 \\
4  & the\_DET           & 0.107 \\
5  & that\_SCONJ        & 0.085 \\
6  & 10\_NUM            & 0.055 \\
7  & suggest\_VERB      & 0.053 \\
8  & than\_ADP          & 0.051 \\
9  & significantly\_ADV & 0.045 \\
10 & show\_VERB         & 0.039 \\
\end{tabular}
\caption{Top-10 wLAS for Falcon3 7B Base.}
\label{tab:wlas_falcon3_7b_base}
\end{table}

\begin{table}[!htbp]
\centering
\small
\setlength{\tabcolsep}{6pt}
\begin{tabular}{r l r}
\hline
\hline
\textbf{Rank} & \textbf{lemma+UPOS} & \textbf{wLAS} \\
\hline
1  & that\_SCONJ      & 0.135 \\
2  & result\_NOUN     & 0.127 \\
3  & be\_AUX          & 0.119 \\
4  & suggest\_VERB    & 0.109 \\
5  & also\_ADV        & 0.078 \\
6  & the\_DET         & 0.065 \\
7  & than\_ADP        & 0.059 \\
8  & may\_AUX         & 0.052 \\
9  & these\_DET       & 0.048 \\
10 & addition\_NOUN   & 0.040 \\
\end{tabular}
\caption{Top-10 wLAS for Gemma 3 4B Base.}
\label{tab:wlas_gemma3_4b_base}
\end{table}

\begin{table}[!htbp]
\centering
\small
\setlength{\tabcolsep}{6pt}
\begin{tabular}{r l r}
\hline
\hline
\textbf{Rank} & \textbf{lemma+UPOS} & \textbf{wLAS} \\
\hline
1  & that\_SCONJ      & 0.148 \\
2  & result\_NOUN     & 0.142 \\
3  & be\_AUX          & 0.125 \\
4  & suggest\_VERB    & 0.116 \\
5  & also\_ADV        & 0.081 \\
6  & the\_DET         & 0.067 \\
7  & than\_ADP        & 0.052 \\
8  & may\_AUX         & 0.041 \\
9  & these\_DET       & 0.040 \\
10 & conclusion\_NOUN & 0.037 \\
\end{tabular}
\caption{Top-10 wLAS for Llama 3.1 8B Base.}
\label{tab:wlas_llama31_8b_base}
\end{table}

\begin{table}[!htbp]
\centering
\small
\setlength{\tabcolsep}{6pt}
\begin{tabular}{r l r}
\hline
\hline
\textbf{Rank} & \textbf{lemma+UPOS} & \textbf{wLAS} \\
\hline
1  & result\_NOUN       & 0.145 \\
2  & that\_SCONJ        & 0.145 \\
3  & be\_AUX            & 0.123 \\
4  & suggest\_VERB      & 0.110 \\
5  & also\_ADV          & 0.082 \\
6  & the\_DET           & 0.071 \\
7  & than\_ADP          & 0.070 \\
8  & significantly\_ADV & 0.055 \\
9  & conclusion\_NOUN   & 0.046 \\
10 & 10\_NUM            & 0.040 \\
\end{tabular}
\caption{Top-10 wLAS for Mistral v0.3 7B Base.}
\label{tab:wlas_mistral_v03_7b_base}
\end{table}

\begin{table}[!htbp]
\centering
\small
\setlength{\tabcolsep}{6pt}
\begin{tabular}{r l r}
\hline
\hline
\textbf{Rank} & \textbf{lemma+UPOS} & \textbf{wLAS} \\
\hline
1  & be\_AUX            & 0.141 \\
2  & also\_ADV          & 0.139 \\
3  & result\_NOUN       & 0.120 \\
4  & the\_DET           & 0.100 \\
5  & that\_SCONJ        & 0.086 \\
6  & suggest\_VERB      & 0.066 \\
7  & than\_ADP          & 0.056 \\
8  & significantly\_ADV & 0.052 \\
9  & a\_DET             & 0.039 \\
10 & 0\_NUM             & 0.033 \\
\end{tabular}
\caption{Top-10 wLAS for OLMo 2 11B Base.}
\label{tab:wlas_olmo2_11b_base}
\end{table}

\begin{table}[!htbp]
\centering
\small
\setlength{\tabcolsep}{6pt}
\begin{tabular}{r l r}
\hline
\hline
\textbf{Rank} & \textbf{lemma+UPOS} & \textbf{wLAS} \\
\hline
1  & that\_SCONJ        & 0.166 \\
2  & result\_NOUN       & 0.146 \\
3  & suggest\_VERB      & 0.122 \\
4  & be\_AUX            & 0.115 \\
5  & than\_ADP          & 0.072 \\
6  & also\_ADV          & 0.068 \\
7  & the\_DET           & 0.067 \\
8  & significantly\_ADV & 0.066 \\
9  & these\_DET         & 0.051 \\
10 & may\_AUX           & 0.048 \\
\end{tabular}
\caption{Top-10 wLAS entries for Yi 1.5 6B Base.}
\label{tab:wlas_yi15_6b_base}
\end{table}

\FloatBarrier

\section*{Bibliographical References}

\vspace{-0.3cm}

\bibliographystyle{lrec2026-natbib}
\bibliography{lrec2026-example}

\end{document}